\documentclass[11pt]{article}
\usepackage{acl-hlt2011}
\usepackage{times}
\usepackage{latexsym}
\usepackage{amsmath}
\usepackage{multirow}
\usepackage{url}
\usepackage{graphics}
\usepackage{graphicx}
\usepackage{subfigure}
\usepackage[utf8]{inputenc}
\usepackage{tikz}
\usepackage{booktabs}
\usepackage{verbatim}
\usepackage{longtable}
\usepackage{multirow}\usepackage[french]{babel}

\title{French parsing enhanced with a word clustering method based on a syntactic lexicon}

\author{Anthony Sigogne \\
  Université Paris-Est, LIGM \\
  {\tt sigogne@univ-mlv.fr} \\\And
  Matthieu Constant \\
  Université Paris-Est, LIGM \\
  {\tt mconstan@univ-mlv.fr} \\\And
  \'Eric Laporte \\
  Université Paris-Est, LIGM \\
  {\tt laporte@univ-mlv.fr} \\}

\date{}

\begin{document}
\maketitle
\begin{abstract}
This article evaluates the integration of data extracted from a French syntactic lexicon, the Lexicon-Grammar \cite{Gross94}, into a probabilistic parser. We show that by applying clustering methods on verbs of the French Treebank \cite{Abeille-etal-2003-building}, we obtain accurate performances on French with a parser based on a Probabilistic Context-Free Grammar \cite{Petrov-etal-2006-learning}.
\end{abstract}

\section{Introduction}
Syntactic lexicons are rich language resources that may contain useful data for parsers like subcategorisation frames, as it provides, for each lexical entry, information about its syntactic behaviors. Many works on probabilistic parsing studied the use of a syntactic lexicon. We can cite Lexical-Functional Grammar [LFG] \cite{Odonovan-etal-2005-large,Schluter-van-genabith-2008-treebank}, Head-Driven Phrase Structure Grammar [HPSG] \cite{10.1007/978-3-540-30211-7_68} and Probabilistic Context-Free Grammars [PCFG] \cite{Briscoe-carroll-1997-automatic,Deoskar-2008-estimation}. The latter has incorporated valence features of verbs to PCFGs and observe slight improvements on global performances. However, the incorporation of syntactic data on part-of-speech tags increases the effect of data sparseness, especially when the PCFG grammar is extracted from a small treebank\footnote{Data sparseness implies the difficulty of estimating probabilities of rare rules extracted from the corpus.}. \cite{Deoskar-2008-estimation} was forced to reestimate parameters of his grammar with an unsupervised algorithm applied on a large raw corpus. In the case of French, this observation can be linked to experiments described in \cite{Crabbe-candito-2008-experiences} where POS tags are augmented with some syntactic functions\footnote{There were 28 original POS tags and each can be combined with one of the 8 syntactic functions.}. Results have shown a huge decrease on performances.\\
The problem of data sparseness for PCFG is also lexical. The richer the morphology of a language, the sparser the lexicons built from a treebank will be for that language. Nevertheless, the effect of lexical data sparseness can be reduced by word clustering algorithms. Inspired by the clustering method of \cite{Koo-etal-2008-simple}, \cite{Candito-crabbe-2009-improving,Candito-etal-2010-statistical} have shown that by replacing each word of the corpus by automatically obtained clusters of words, they can improve a PCFG parser on French. They also created two other clustering methods. A first method consists in a step of \textit{desinflection} that removes some inflexional marks of words which are considered less important for parsing. Another method consists in replacing each word by the combination of its POS tag and lemma. Both methods improve significantly performances.\\
In this article, we propose a clustering method based on data extracted from a syntactic lexicon, the Lexicon-Grammar. This lexicon offers a classification of lexical items into tables, each table being identifiable by its unique identifier. A lexical item is a lemmatized form which can be present in one or more tables depending on its meaning and its syntactic behaviors. The clustering method consists in replacing a verb by the combination of its POS tag and its tables identifiers. The goal of this article is to show that a syntactic lexicon, like the Lexicon-Grammar, which is not originally developed for parsing algorithms, is able to improve performances of a probabilistic parser.\\
In section 2 and 3, we describe the probabilistic parser and the treebank, namely the French Treebank, used in our experiments. In section 4, we describe more precisely previous work on clustering methods. Section 5 introduces the Lexicon-Grammar. We detail information contained in this lexicon that can be used in the parsing process. Then, in section 6, we present methods to integrate this information into parsers and, in section 7, we describe our experiments and discuss the obtained results.

\section{Non-lexicalized PCFG parser}
The probabilistic parser, used into our experiments, is the Berkeley Parser\footnote{The Berkeley Parser is freely available at http://code.google.com/p/berkeleyparser/} (called BKY thereafter) \cite{Petrov-etal-2006-learning}. This parser is based on a PCFG model which is non-lexicalized. The main problem of non-lexicalized context-free grammars is that nonterminal symbols encode too general information which weakly discriminates syntactic ambiguities. The benefit of BKY is to try to solve the problem by generating a grammar containing complex symbols. It follows the principle of latent annotations introduced by \cite{Matsuzaki-etal-2005-probabilistic}. It consists in iteratively creating several grammars, which have a tagset increasingly complex. For each iteration, a symbol of the grammar is splitted in several symbols according to the different syntactic behaviors of the symbol that occur into a treebank. Parameters of the latent grammar are estimated with an algorithm based on Expectation-Maximisation (EM). In the case of French, \cite{Seddah-etal-2009-adaptation} have shown that BKY produces \textit{state-of-the-art} performances.

\section{French Treebank}
For our experiments, we used the French Treebank\footnote{The French Treebank is freely available under licence at http://www.llf.cnrs.fr/Gens/Abeille/French-Treebank-fr.php} \cite{Abeille-etal-2003-building} [FTB]. It is composed of articles from the newspaper \textit{Le Monde} where each sentence is annotated with a constituent tree. Currently, most of papers about parsing of French use a specific variant of the FTB, namely the FTB-UC described for the first time in \cite{Candito-crabbe-2009-improving}. It is a partially corrected version of the FTB which contains 12351 sentences and 350931 tokens. This version is smaller\footnote{The original FTB contains 20,648 sentences and 580,945 tokens.} and has specific characteristics. First, the tagset takes into account the rich original annotation containing morphological and syntactic information. It results in a tagset of 28 part-of-speech tags. Some compounds with regular syntax schemas are undone into phrases containing simple words. Remaining compounds are merged into a single token, whose components are separated with an underscore.

\section{Previous work on word clustering}
\label{previous}
Numerous works used clustering methods in order to reduce the size of the corpus lexicon and therefore reducing the impact of lexical data sparseness on treebank grammars. A method, described in \cite{Candito-seddah-2010-parsing} and called \textit{CatLemma}, consists in replacing a word by the combination of its POS tag and its lemma. In the case of a raw text to analyze (notably during evaluations), they used a statistical tagger in order to assign to each word both POS tag and lemma\footnote{They used the tagger MORFETTE \cite{Chrupala-etal-2008-learning,Seddah-etal-2010-lemmatization} which is based on two statistical models, one for tagging and the other for lemmatization. Both models were trained thanks to the \textit{Average Sequence Perceptron} algorithm.}.\\
Instead of reducing each word to the lemmatized form, \cite{Candito-crabbe-2009-improving,Candito-seddah-2010-parsing} have done a morphological clustering, called \textit{desinflection} [DFL], which consists in removing morphological marks that are \textit{less important} for determining syntactic projections in constituents. The mood of verbs is, for example, very helpful. On the other hand, some marks, like gender or number for nouns or the person of verbs, are not so crucial. Moreover, original ambiguities on words are kept in order to delegate the task of POS tags desambiguation to the parser. This algorithm is done with the help of a morpho-syntactic lexicon.\\
The last clustering method, called \textit{Clust}, consists in replacing each word by a cluster id. Cluster ids are automatically obtained thanks to an unsupervised statistical algorithm \cite{Brown-etal-1992-class} applied on a large raw corpus. They are computed by taking account of co-occurrence information of words. The main advantage of this method is the possibility of combining it to \textit{DFL} or \textit{CatLemma}. First, the raw corpus is preprocessed with one of these two methods and then, clusters are computed on this modified corpus. Currently, this method permits to obtain the best results on the FTB-UC.

\section{Lexicon-Grammar}
\begin{figure}[h]
    \centering
    \includegraphics[scale=0.3]{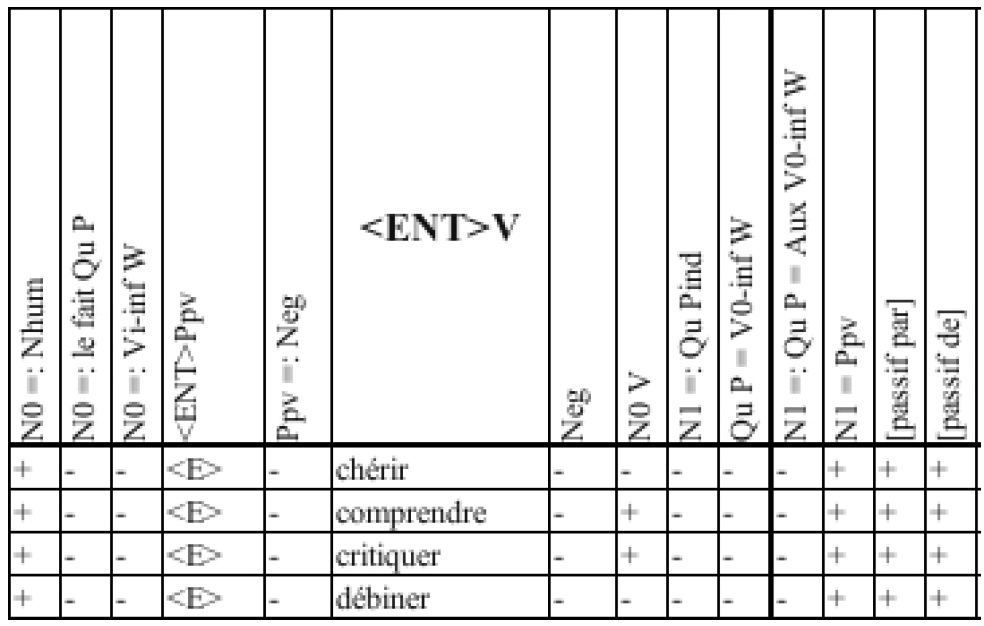}
    \caption{Sample of verb table \textit{12}.}
    \label{table1}
\end{figure}

\begin{figure}[h]
    \centering
    \includegraphics[scale=0.35]{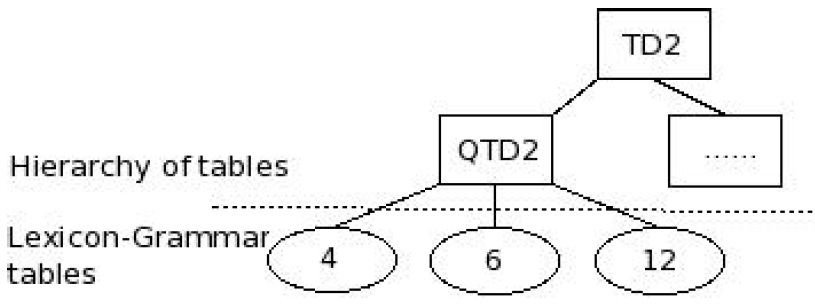}
    \caption{Sample of the hierarchy of verb tables.}
    \label{hierarchy}
\end{figure}

The Lexicon-Grammar [LG] is the richest source of syntactic and lexical information for French\footnote{We can also cite lexicons like LVF \cite{Dubois97}, Dicovalence \cite{Eynde03} and Lefff \cite{Sagot-2010-lefff}.} that focuses not only on verbs but also on verbal nouns, adjectives, adverbs and frozen (or fixed) sentences. Its evelopment started in the 70's by Maurice Gross and his team \cite{Gross94}. It is a syntactic lexicon represented in the form of tables. Each table encodes lexical items of a particular category sharing several syntactic properties (e.g. subcategorization information). A lexical item is a lemmatized form which can be present in one or more tables depending on its meaning and its syntactic properties. Each table row corresponds to a lexical item and a column corresponds to a property (e.g. syntactic constructions, argument distribution, and so on). A cell encodes whether a lexical item accepts a given property. Figure~\ref{table1} shows a sample of verb table \textit{12}. In this table, we can see that the verb \textit{chérir} (\textit{to cherish}) accepts a human subject (pointed out by a $+$ in the property \textit{N0 =: Nhum}) but this verb cannot be intransitive (pointed out by a $-$ in the property \textit{N0 V}).
Recently, these tables have been made consistent and explicit \cite{Tolone11} in order to be exploitable for NLP. They also have been transformed in a XML-structured format \cite{Constant2008generictoolgeneratelexicon}\footnote{These resources are freely available at http://infolingu.univ-mlv.fr/}. Each lexical entry is associated with its table identifier, its possible arguments and its syntactic constructions.\\
For the verbs, we manually constructed a hierarchy of the tables on several levels. Each level contains classes which group LG tables which may not share all their defining properties but have a relatively similar syntactic behavior. Figure~\ref{hierarchy} shows a sample of the hierarchy. The tables \textit{4, 6} and \textit{12} are grouped into a class called \textit{QTD2} (transitive sentence with two arguments and sentential complements). Then, this class is grouped with other classes at the superior level of the hierarchy to form a class called \textit{TD2} (transitive sentence with two arguments).
The characteristics of each level are given in the Table~\ref{reg1b} (level 0 represents the set of tables of the LG). We can state that there are 5,923 distinct verbal forms for 13,862 resulting entries in tables of verbs\footnote{Note that 3,121 verb forms (3,195 entries) are unambiguous. This means that all their entries occur in a single table.}. The column \textit{\#classes} specifies the number of distinct classes. The columns \textit{AVG\_1} and \textit{AVG\_2} respectively indicate the average number of entries per class and the average number of classes per distinct verbal form.
\begin{table}[h]
    \footnotesize
    \centering
    \begin{tabular}{|l|l|l|l|}
        \hline Level & \#classes & AVG\_1 & AVG\_2\\
        \hline 0 & 67 & 207 & 2.15\\
        \hline 1 & 13 & 1,066 & 1.82\\
        \hline 2 & 10 & 1,386 & 1.75\\
        \hline 3 & 4 & 3,465 & 1.44\\
        \hline
    \end{tabular}
	\caption{\small Characteristics of the hierarchy of verb tables.}
    \label{reg1b}
\end{table}\\
The hierarchy of tables has the advantage of reducing the number of classes associated with each verb of the tables. We will see that this ambiguity reduction is crucial in our experiments.

\section{Word clustering based on the Lexicon-Grammar}
The LG contains a lot of useful information that could be used into the parsing process. But such information is not easily manipulable. We will focus on table identifiers of the verb entries which are important hints about their syntactic behaviors. For example, the table~\textit{31R} indicates that all verbs belonging to this table are intransitive. Therefore, we followed the principle of the clustering method \textit{CatLemma}, except that here, we replace each verb of a text by the combination of its POS tag and its table ids associated with this verb in the LG tables\footnote{Verbs that are not in the LG remain unchanged.}. We will call this experiment \textit{TableClust} thereafter. For instance, the verb \textit{chérir} (to cherish) belongs to the table \textit{12}. Therefore, the induced word is \textit{\#tag\_12}, where \textit{\#tag} is the POS tag associated with the verb. For an ambiguous verb like \textit{sanctionner} (to punish), belonging to two tables \textit{6} and \textit{12}, the induced word is \textit{\#tag\_6\_12}.\\
Then, we have done variants of the previous experiment by taking the hierarchy of verb tables into account. This hierarchy is used to obtain clusters of verbs increasingly coarse as the hierarchy level increases, and at the same time, the size of the corpus lexicon is also increasingly reduced. Identifiers combined to the tag depend on the verb and the specific level in the hierarchy. For example, the verb \textit{sanctionner}, belonging to tables \textit{6} and \textit{12}, is replaced by \textit{\#tag\_QTD2} at level 1. In the case of ambiguous verbs, for a given level in the hierarchy, identifiers are all classes the verb belongs to. This experiment will be called \textit{LexClust} thereafter.\\
As for clustering method \textit{CatLemma}, we need a Part-Of-Speech tagger in order to assign a tag and a lemma to each verb of a text (table ids can be determined from the lemma). We made the choice to use \textit{MElt} \cite{Denis-sagot-2009-coupling} which is one of the best taggers for French. Lemmatization process is done with a French dictionary, the Dela \cite{CourtoisSilb1990}, and some heuristics in the case of ambiguities.

\section{Experiments and results}
\subsection{Evaluation metrics}
As the FTB-UC is a small corpus, we used a \textit{cross-validation} procedure for evaluations. This method consists in splitting the corpus into \textit{p} equal parts, then we compute training on \textit{p-1} parts and evaluations on the remaining part. We can iterate this process \textit{p} times. This allows us to calculate an average score for a sample as large as the initial corpus. In our case, we set the parameter \textit{p} to 10. Results on evaluation parts are reported using the standard protocol called PARSEVAL \cite{10.3115/112405.112467} for all sentences. The labeled F-Measure [F1] takes into account the bracketing and labeling of nodes. We also use the unlabeled and labeled attachement scores [UAS, LAS] which evaluate the quality of unlabeled and labeled dependencies between words of the sentence\footnote{These scores are computed by automatically converting constituent trees into dependency trees. The conversion procedure is made with the \textit{Bonsaï} software, available at http://alpage.inria.fr/statgram/frdep/fr\_stat\_dep\_parsing.html.}. Punctuation tokens are ignored in all metrics.

\subsection{Berkeley parser settings}
We used a modified version of BKY enhanced for tagging unknown and rare French words \cite{Crabbe-candito-2008-experiences}\footnote{Available in the \textit{Bonsaï} package.}. We can notice that BKY uses two sets of sentences at training, a learning set and a validation set for optimizing the grammar parameters. As in \cite{Candito-etal-2010-statistical}, we used 2\% of each training part as a validation set and the remaining 98\% as a learning set. The number of split and merge cycles was set to 5.

\subsection{Clustering methods}
We have evaluated the impact of clustering methods \textit{TableClust} and \textit{LexClust} on the FTB-UC. For both methods, verbal forms of each training part are replaced by the corresponding cluster and, in order to do it on the evaluation part, we use Melt and some heuristics. So as to compare our results with previous work on word clustering, we have reported results of two clustering methods described in section~\ref{previous}, \textit{DFL} and \textit{DFL+Clust} ($Clust$ is applied on a text that contains $desinflected$ words).

\subsection{Evaluations}
The experimental results are shown in the Table~\ref{reg2b}\footnote{All experiments have a tagging accuracy of about 97\%.}. The column \textit{\#lexicon} represents the size of the FTB-UC lexicon according to word clustering methods. In the case of the method \textit{LexClust}, we varied the level of the verbs hierarchy used.
\setlength{\doublerulesep}{\arrayrulewidth}
\begin{table}[h]
    \footnotesize
    \centering
    \begin{tabular}{|l|l|l|l|l|l|}
	\hline Method & \#lexicon & F1 & UAS & LAS & F1$<$40\\
	\hline Baseline & 27,143 & 83.82 & 89.43 & 85.85 & 86.12\\
	\hline\hline DFL & 20,127 & 84.57 & 89.91 & 86.36 & 86.80\\
	\hline DFL+Clust & 1,987 & 85.22 & 90.26 & 86.70 & 87.39 \\
	\hline\hline TableClust & 24,743 & 84.11 & 89.67 & 86.10 & 86.53\\
	\hline LexClust~1 & 22,318 & 84.33 & 89.77 & 86.22 & 86.62\\
	\hline LexClust~2 & 21,833 & 84.44 & 89.87 & 86.32 & 86.76\\
	\hline LexClust~3 & 20,556 & 84.26 & 89.64 & 86.10 & 86.57\\
    \hline\hline Tag & 20478 & 84.11 & 89.58 & 86.00 & 86.40\\
	\hline TagLemma & 24722 & 83.87 & 89.51 & 85.91 & 86.26\\
	\hline
    \end{tabular}
	\caption{\small Results from cross-validation evaluation according to clustering methods.}
    \label{reg2b}
\end{table}
The method \textit{TableClust} slightly improves performances compared with the baseline. Nevertheless, using levels of the hierarchy of verb tables through \textit{LexClust} increases results while considerably reducing the size of the corpus lexicon. We obtain the best results with the level 2 of the hierarchy. These performances are almost identical to those of \textit{DFL}, despite the fact that we only modify verbal forms while \textit{DFL} alters all inflected forms regardless of grammatical categories. However, \textit{DFL+Clust} has high scores and is significantly better than \textit{LexClust}. As of this writing, we tried some combination of methods \textit{LexClust} and \textit{Clust} but we observed that both methods are not easily mergeable.\\
The impact of \textit{TableClust} and \textit{LexClust} on a new text is strongly influenced by the quality of the tagging produced by \textit{Melt}. For evaluating this impact, we computed \textit{Gold} experiments for both clustering method. Each verb of evaluation parts, present in the LG tables, is replaced by correct tag and table ids. We observed a gap of almost 0.5\% for both tagging and F1. For instance, on the first evaluation part, \textit{Melt} has high but not perfect scores, with a precision of 98.2\% and a recall of 97.2\%, for a total of 165 errors\footnote{We can compute precision and recall scores because sometimes \textit{Melt} wrongly identifies a word as a verb or miss a verb.}. About lemmatization, we have a perfect score of 100\%.\\
Our approach is based on the combination of tags and table ids contained in the syntactic lexicon. In order to validate this approach, we have done two other experiments. A first one, called \textit{Tag}, consists in replacing each verbal lemma by its verbal tag only. The second one, called \textit{TagLemma}, consists in the combination of the tag and the lemma. Results are reported in the Table~\ref{reg2b}. As for \textit{TableClust} and \textit{LexClust}, we replace only verbal forms that are present in the LG tables. We can see that \textit{Tag} has equal performances to \textit{TableClust}. Therefore, original table ids combined with tags are useless. Maybe, the number of clusters is too high and consequently, the size of the corpus lexicon is still too large. However, \textit{LexClust} is better than \textit{Tag}. About \textit{TagLemma}, results are almost identical to the baseline. According to these observations, we can say that verbal clusters created with our method \textit{LexClust} are relevant and useful for a parser like BKY.\\
We have indicated in Table~\ref{err}, the top most F1 absolute gains according to phrase labels, for our best clustering method \textit{LexClust} with level 2 of the hierarchy. For each phrase, the column called \textit{Gain} indicates the average F1 absolute gain in comparison to the baseline F1 for this phrase, and \textit{prop.} is the proportion of the phrase in the whole corpus. 
\begin{table}[h]
    \footnotesize
    \centering
    \begin{tabular}{|l|l|l|}
        \hline Phrase label & Meaning & Gain (prop.)\\
	   \hline VPpart & participial phrase & 4,4\% (2\%)\\
	   \hline Srel   & relative clause    & 1,6\% (1\%)\\
	   \hline VN     & verbal nucleus      & 1.1\% (11\%)\\
        \hline VPinf  & infinitive phrase  & 0.9\% (0.4\%)\\
	   \hline AdP    & adverbial phrase   & 0.9\% (3\%)\\	
        \hline
    \end{tabular}
	\caption{\small Top most F1 absolute gains according to phrases.}
    \label{err}
\end{table}
We can see that three of the five best corrected phrases relate to verbal phrases (plus one if we consider that AdP is linked to a verbal phrase). Therefore, the integration of syntactic data into a clustering algorithm of verbs improves the recognition of verbal phrases.
%\footnote{Note that \textit{LexClust} with other levels of the hierarchy gives the same best corrected phrases.}

\section{Conclusion and future work}
In this article, we have shown that by using information on verbs from a syntactic lexicon, like the Lexicon-Grammar, we are able to improve performances of a statistical parser based on a PCFG grammar. In the near future, we plan to reproduce experiments with other grammatical categories.
%, like nouns or adjectives.

\bibliographystyle{acl}
\bibliography{biblio} 

\end{document}